\documentclass[letterpaper, 10 pt, conference]{IEEEtran}  % Comment this line out if you need a4paper

\pdfoutput=1

\IEEEoverridecommandlockouts                              % This command is only needed if 
\usepackage{graphics} % for pdf, bitmapped graphics files
\usepackage{epsfig} % for postscript graphics files
\usepackage{mathptmx} % assumes new font selection scheme installed
\usepackage{times} % assumes new font selection scheme installed
\usepackage{amsmath} % assumes amsmath package installed
\usepackage{amssymb}  % assumes amsmath package installed
\usepackage{tikz}

\usepackage{hyperref}

\title{\LARGE \bf
Learning Bipedal Robot Locomotion from Human Movement
}

\newcommand\copyrighttext{%
  \footnotesize \textcopyright 2021 IEEE. Personal use of this material is permitted. Permission from IEEE must be obtained for all other uses, in any current or future media, including reprinting/republishing this material for advertising or promotional purposes, creating new collective works, for resale or redistribution to servers or lists, or reuse of any copyrighted component of this work in other works.
  }
\newcommand\copyrightnotice{%
\begin{tikzpicture}[remember picture,overlay]
\node[anchor=south,yshift=10pt] at (current page.south) {\fbox{\parbox{\dimexpr\textwidth-\fboxsep-\fboxrule\relax}{\copyrighttext}}};
\end{tikzpicture}%
}

\author{Michael Taylor$^1$, Sergey Bashkirov$^1$, Javier Fernandez Rico$^1$, Ike Toriyama$^1$ \\ Naoyuki Miyada$^1$, Hideki Yanagisawa$^1$, Kensaku Ishizuka$^1$ \\ $^1$Sony Interactive Entertainment
}

\begin{document}

\maketitle
\copyrightnotice
\thispagestyle{empty}
\pagestyle{empty}

%%%%%%%%%%%%%%%%%%%%%%%%%%%%%%%%%%%%%%%%%%%%%%%%%%%%%%%%%%%%%%%%%%%%%%%%%%%%%%%%
\begin{abstract}

  Teaching an anthropomorphic robot from human example offers the opportunity to impart humanlike qualities on its movement. In this work we present a reinforcement learning based method for teaching a real world bipedal robot to perform movements directly from human motion capture data. Our method seamlessly transitions from training in a simulation environment to executing on a physical robot without requiring any real world training iterations or offline steps. To overcome the disparity in joint configurations between the robot and the motion capture actor, our method incorporates motion re-targeting into the training process. Domain randomization techniques are used to compensate for the differences between the simulated and physical systems. We demonstrate our method on an internally developed humanoid robot with movements ranging from a dynamic walk cycle to complex balancing and waving. Our controller preserves the style imparted by the motion capture data and exhibits graceful failure modes resulting in safe operation for the robot. This work was performed for research purposes only.

\end{abstract}

%%%%%%%%%%%%%%%%%%%%%%%%%%%%%%%%%%%%%%%%%%%%%%%%%%%%%%%%%%%%%%%%%%%%%%%%%%%%%%%%
\section{INTRODUCTION}
  When a robot possesses a humanoid form factor, the expectation is conveyed that it will move in a human like manner. Classic control theory approaches struggle to achieve credible results. In general they exhibit unnatural motions that appear mechanical in nature. To bridge this gap we employ reinforcement learning (RL) \cite{reinforcement_learning} to train a neural network to control a humanoid robot using human motion capture data as a reference. While robust motion imitation solutions based on RL exist in simulated environments \cite{deep_mimic}, translating the success seen in simulation to the real world has remained a challenging problem.

  Issues encountered when transferring from simulation to the real world may include the following:
\begin{itemize}
  \item Real world actuators exhibit a variety of imperfections including jitter, bias and motor backlash. 
  \item On board sensors suffer from bias and noise.
  \item The control system must respond to multiple noisy sensor inputs simultaneously while avoiding over-reacting to any single sensor reading.
  \item Smoothing techniques applied to reduce the sensor spikes in general come at the cost of responsiveness to the sensor inputs which is vital to maintaining stability.
  \item Partial observability limits real world observations to sensors readings, whereas in simulation the full dynamical state of the system is available.
  \item Physics simulations make use of simplified models when calculating forces such as friction \cite{gazebo}.
\end{itemize}
  Such limitations make domain adaptation from a simulation environment to a real world environment difficult.

  We propose the following techniques to overcome these challenges:
\begin{itemize}
  \item We implement a control scheme that treats the neural network output as joint position derivatives and then integrate the position derivatives into target joint positions.
	\item We describe a progress based reward function that incorporates animation re-targeting into the training process.
  \item We employ a domain randomization scheme that relies on simulated motor backlash, randomized friction, and randomized Young's modulus to enable the controller to transfer between simulation and the real world.
\end{itemize}
  We apply the controller trained in simulation to a real world humanoid robot and demonstrate a series of complex motions learned directly from human motion capture data. We show our method preserves the style imparted by the motion capture data when transferred to the real robot.

\section{RELATED WORK}
  A large body of literature has been published over the last decade on applying RL to character control \cite{trajectory, deep_loco, deep_mimic}. An overview of research in the field from previous decades can be found in \cite{overview}. A similar body of work exists on applying RL to robotic control \cite{quadrupedal, simple_biped, data_biped}.  We highlight a select set of recent works that we feel are most relevant to our own.
  
  Peng et al., 2018 \cite{deep_mimic} employs RL to learn a control policy that enables a simulated character to imitate human movements. They limit their focus to simulation and require the kinematic structure of the trained character to match that of the reference character. We present a number of design choices that in aggregate allow our solution to map between characters with different morphologies as well as transfer from simulation to the real world.
 
  Xie et al., 2018 \cite{cassie_2018} uses RL to train a robust walking controller in simulation that imitates a reference motion. In contrast to using a single reference trajectory consisting of two symmetrical steps, we train on a diverse set of non-symmetric motion clips lifted from human motion capture data. Furthermore, our targets do not provide a stable reference motion adapted to the robot's form factor. While their policy learns how to augment the reference motion, we limit our use of the reference motion to the training phase. They demonstrate their results using a non-humanoid robot consisting only of lower body limbs, while we demonstrate our system using a full-body humanoid robot.
  
  Xie et al., 2019 \cite{cassie_2019} introduces the data collection technique deterministic action stochastic state (DASS). The data consists of a set of deterministic decision tuples $(s, a)$ gathered from one or more expert policies. The DASS samples are combined with conventional RL-based policy-gradient samples when performing network updates. This allows them to forego domain randomization when transferring from simulation to the real world, which they demonstrate on a single surface type. In contrast, we incorporate domain randomization techniques that allow our method to generalize across multiple surfaces.
 
  Peng et al., 2020 \cite{animal_imitation} presents an imitation learning system that enables quadruped robots to learn agile locomotion skills by imitating real-world animals. They use an information bottleneck for domain adaptation that allows the control policy to transfer from simulation to the real world. Environment information is passed through the information bottleneck, which constrains the neural network's view of the environment. By adjusting the restrictiveness of the bottleneck, they are able to control the amount of generalization the solution exhibits. This approach requires multiple real world iterations in order to infer a suitable set of parameters for executing a skill in the real world. In contrast, our method is able to transfer from simulation to the physical world without interacting with the real world. By bypassing choosing an environment latent space representation, we avoid specializing our policy towards a specific environment.

\section{DYNAMICS MODEL}
  We consider a robot to be a humanlike mechanism controlled via actuators.  Our goal is to generate a series of commands for the actuators that guide the robot through a target motion.  We utilize a dynamics simulation environment (ODE) \cite{ode, gazebo} to model the mechanical response of the actuators to the control input.  Each body part (link) of the robot is represented as one or more rigid bodies. The robot's links are connected via revolute joints.  In the physical world, all robot joints are driven by servo controllers that apply torques to the joints when required.  We use the proportional derivative (PD) regulator model of a servo controller to represent the controllers in simulation.

  Our robot is equipped with several sensors through which it perceives its state in relation to the environment.  These are modeled as virtual sensors in simulation. The list of sensors includes: 
\begin{itemize}
	\item An Inertia Measurement Unit (IMU) that measures the acceleration and angular velocity of the torso.
	\item Joint angle sensors that measure the joint rotations.
	\item Foot pressure sensors that measure surface reaction forces. 
	\item An internal clock that measures time.
\end{itemize}
In simulation each sensor reading is the superposition of the idealized reading combined with simulated bias and noise.

\section{BACKGROUND}
  We pose our problem as a standard reinforcement learning task where an agent learns to maximize the reward received from interacting with an environment. We label our robot as the agent and our physics simulation as the environment. We define the state $s$ as the set of all sensor readings the robot uses to determine its configuration at each time step $t$. Our goal is to learn a policy $\pi$ that maps each state to an action $a$ consisting of a set of updated joint position derivatives. The policy is defined as a probability distribution over the available actions for a given state $\pi(a_t|s_t)$. We consider our environment as a Markov Decision Process, where the state transition probability of reaching state $s_{t+1}$ from state $s_t$ by taking action $a_t$ is described by the function \eqref{eq:transition_definition}:
  \begin{align}
  \label{eq:transition_definition}
    & T(s,a,s') = \mathbb{P}[S_{t+1}=s' | S_t=s,A_t=a].
  \end{align}
We define $r$ as the reward received when transitioning from state $s_t$ to state $s_{t+1}$ by taking action $a_t$ \eqref{eq:reward_definition}.
  \begin{align}
  \label{eq:reward_definition}
    & r_t = R(s_t, a_t, s_{t+1})
  \end{align}
  
  Policy gradient methods are a common class of algorithms used for continuous action problems \cite{simple_grad, policy_gradient}. They attempt to learn a policy by iteratively improving the policy using the gradient of the objective function with respect to the policy. The gradient is calculated as the expectation of the log-likelihood of an action under the policy at a given state multiplied by the return received by the agent when performing that action at the state \eqref{eq:policy_gradient}. Trajectories are generated by an agent executing a series of actions sampled from a policy. The Q-function $Q^\pi(s,a)$ estimates the return, the expected cumulative discounted future reward across multiple time steps, from taking action $a$ sampled from policy $\pi$ at state $s$.
  \begin{align}
  \label{eq:policy_gradient}
    & \nabla_\theta J(\theta)  =  \mathbb{E}_{s, a \sim \pi(s, a)}[Q^\pi(s, a) \nabla_\theta \ln \pi_\theta(a \vert s)]
  \end{align}
  
  We train our policies with an actor-critic algorithm \cite{actor_critic}. The actor is a policy network and the critic a value network. The value from the critic is used to calculate an advantage \eqref{eq:advantage}, which is defined as the value of taking an action from a state without regard to the underlying value of the state \cite{advantage}. The advantage is used in place of the return when calculating the policy gradient:
  \begin{align}
  \label{eq:advantage}
    & A(s, a) = Q(s, a) - V(s).
  \end{align}

\section{IMPLEMENTATION}
  The policy is represented by a neural network and the target motion is specified in the form of a motion capture clip. We make use of the PPO algorithm \cite{ppo} together with Generalized Advantage Estimation \cite{gae} to train the policy.

  A problem exists if the agent differs in either degrees of freedom (DOF) or proportion from the morphology of the motion capture actor. Both scenarios prevent the use of joint position differences as a reward. To resolve the disparity in proportions, we match the link orientations between the agent and target. To handle the difference in the number of DOF, we map a subset of the agent links to the target links. This permits the unmapped links to freely rotate in the way that best reduces the orientation error of the mapped links.

  In the specific case of the agent possessing a lower number of DOF than the target, a subset of the agent links will be unable to match the orientations of the target links. Typically, this is the case for joints where a human has three DOF and a robot only two. To overcome this we make use of two separate distance metrics, the quaternion distance between two links and the angle between two link unit axes. We define the link unit axis as a unit vector in the link's local reference frame.

  For the quaternion distance metric the following procedure is used. Let $q_a$ and $q_t$ be the agent and target link orientation quaternions. Then
  \begin{align}
    \label{eq:quat_delta}
    \Delta q = q_a^{*} * q_t
  \end{align}
  \begin{align}
    \label{eq:quat_dist}
    d_{link} = 2 * sin^{-1}( \Delta q_x^2 + \Delta q_y^2 + \Delta q_z^2 )
  \end{align}
  where $d_{link}$ is the quaternion distance between link orientations.
  
  The angle between link axes is computed in the following way. Let $\vec{e_a}$ and $\vec{e_t}$ be the agent and target link unit axes converted to the world reference frame. Then the axis distance is:
  \begin{align}
    \label{eq:axis_dist}
    d_{axis} = cos^{-1}( e_{a,x}*e_{t,x} + e_{a,y}*e_{t,y} + e_{a,z}*e_{t,z} )
  \end{align}
  Using the introduced unit axis distance allows us to map links where there is an insufficient number of DOF.

\section{CONTROL SCHEME}
  There are three commonly used control parameterizations for articulated robots: position control, velocity control, and torque control.  Previous studies have demonstrated neural networks perform best using position control when driving a character for motion imitation \cite{action_space}.

\begin{figure}
    \centering
    \vspace*{1.5mm}
    \includegraphics[width=7cm]{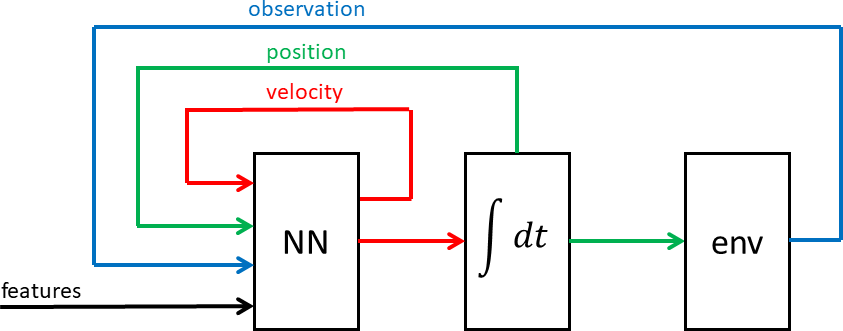}
    \vspace*{-1.5mm}
    \caption{Single integrator. This schema describes the neural network input, output, and feedback loops.}
    \label{fig:single_integrator}
  \end{figure}

  In this work we employ a hybrid scheme by having the neural network output target joint velocities, which for generality we label as position derivatives. The derivatives $v$ are integrated into joint positions according to the formula $\int v dt$ before being applied directly to either the PD controllers in simulation or to the actuators of a real robot. The output of the integrator block is also used as a feedback signal by routing it into the neural network as input (Fig. \ref{fig:single_integrator}).
  
\begin{figure}
    \centering
    \vspace*{1.5mm}
    \includegraphics[width=8.6cm]{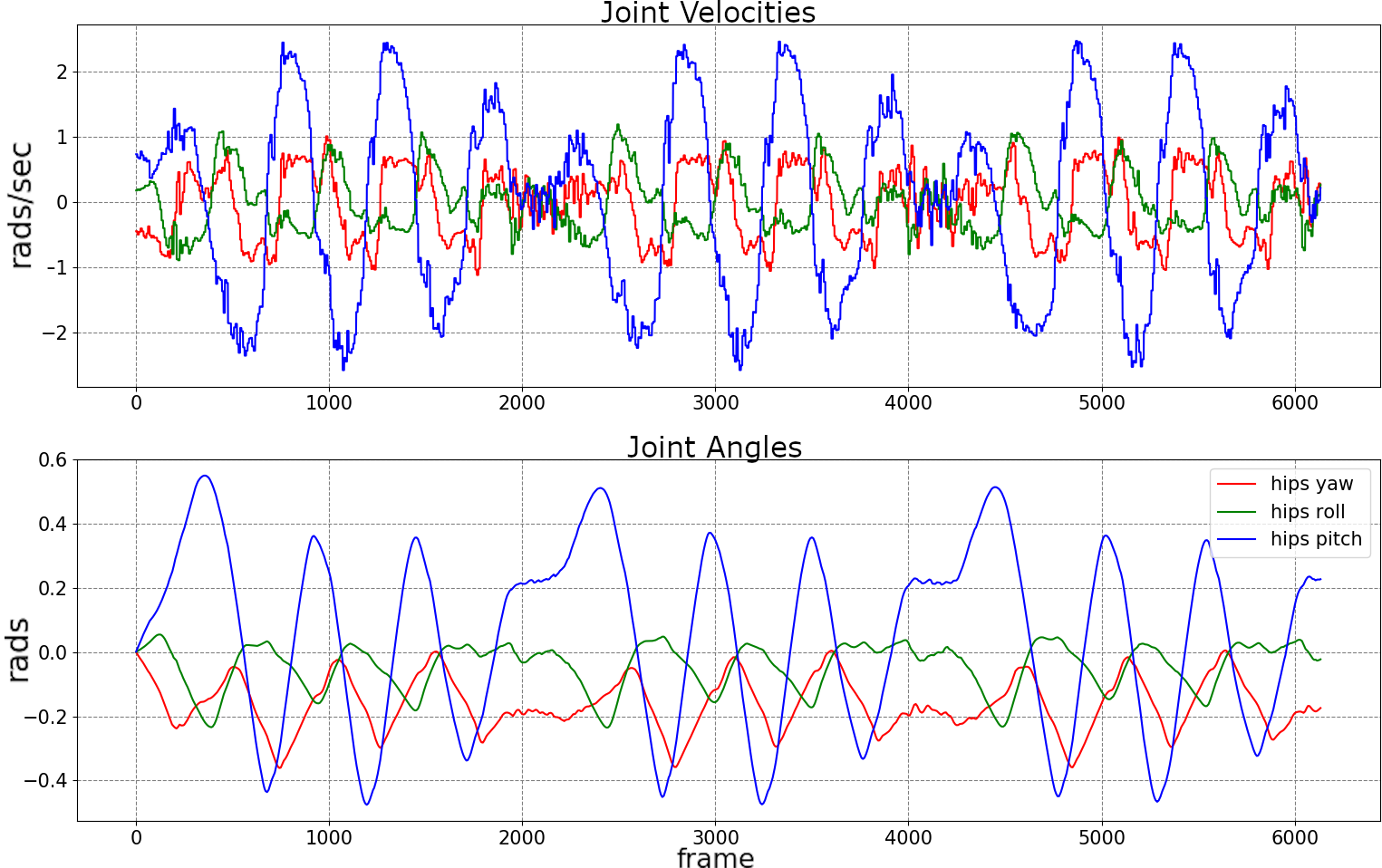}
    \vspace*{-6.5mm}
    \caption{Right hip joint yaw, pitch, and roll position derivatives and their integrated positions over the course of the walk motion.}
    \label{fig:integral_walk}
\end{figure}

  The integration step has several significant properties. It suppresses motor jitter in simulation and the real world to visually unobservable levels. It smooths out the robot's reaction to noisy sensors and sensor spikes. It moderates the robot's movement when the network input enters out-of-distribution areas of the state space during failure scenarios. Fig. \ref{fig:integral_walk} shows how the rapidly changing position derivatives are integrated into smooth positions for the walk motion.
  
  Additional experiments were performed by double integrating the network output. We find that double integration is unable to accurately reproduce movements, likely due to requiring complicated acceleration profiles (see supplementary video).
  
\section{NEURAL NETWORK ARCHITECTURE}
  We split our policy and value functions into separate networks with no shared weights (Fig. \ref{fig:neural_network}).  Both networks consist of three layers.  Each of these layers contain 128 neurons and use softsign as their activation function.  The network input (observation) is normalized using the running mean and standard deviation. The input consists of the following features:

\begin{figure}
  \centering
  \vspace*{-1mm}
  \includegraphics[width=8cm]{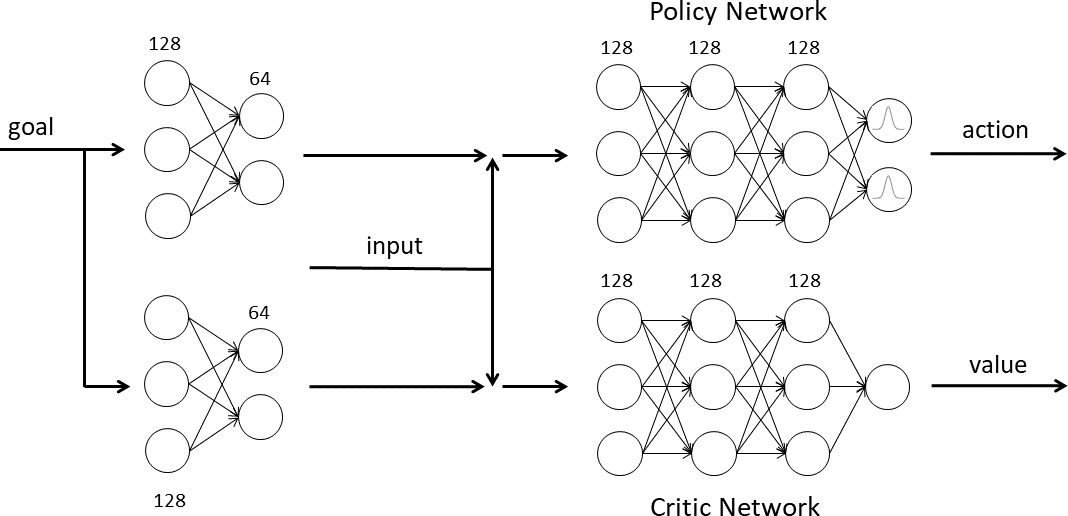}
  \vspace*{-2mm}
  \caption{Policy and value networks.}
  \label{fig:neural_network}
\end{figure}

\begin{itemize}
  \item goal orientations
  \item joint sensor readings
  \item action at previous time step
  \item actuator inputs at previous time step
  \item gravity vector in local reference frame
  \item accelerometer reading
  \item gyro reading
  \item left and right foot pressure sensor readings
\end{itemize}

  The goal orientations are represented in axis-angle form and encoded into a latent representation using two layers of 128 and 64 neurons with leaky ReLU activation functions.  The action specifies the set of joint position derivatives output by the neural network. The actuator inputs indicate the updated joint positions calculated by integrating the position derivatives. Feeding the action and actuator inputs from the previous time step into the networks introduces a feedback signal.

  The policy network specifies a Gaussian distribution over each output represented by a mean and variance $\pi(a|o) = N(\mu(o), \sigma)$.  The means are observation-dependent, while the variances are represented by a vector of values that are independent of the observation. The value network outputs a single scalar value.

  Exploration occurs during training by sampling the policy network output from the learned Gaussian distributions.  Sampling in this manner introduces jitter during training that makes learning difficult as it induces falling. To alleviate the jitter, in addition to our integration scheme (Fig. \ref{fig:single_integrator}) we incorporate a technique from \cite{deep_mimic}. Instead of sampling random actions from the Gaussian distribution at each time step, with fixed probability $\epsilon$ a random action is sampled from the policy and with probability $1-\epsilon$ the agent executes a deterministic action specified by the mean of the Gaussian. Critically, when adopting this scheme updates are performed using only samples where exploration noise is applied.

\section{REWARD FUNCION}
  The reward function consists of the following four terms:
  \begin{align} \label{eq:reward}
    & r_{total} = r_{link} + (-r_{collision}) + r_{ground} + (-r_{limit}).
  \end{align}
The $r_{link}$ term aligns the agent link orientations with the target actor link orientations. We subtract the $L_{2}$ quaternion distance between the agent link and target link at time step $t$ from the $L_{2}$ quaternion distance at time step $t-1$.  This provides a differential error that rewards the link for moving in the correct direction while penalizing it for moving in the wrong direction. The error between the agent pose and target pose is a weighted sum of the individual link orientation errors. If we consider the arms and legs of the robot as link chains, the weights $w_{d}$ are chosen such that the first link in each chain is weighted higher than the last link. This pushes the system to focus on aligning the root links before the end links during training. We use the same differential approach for the link velocities, but only mix a small $w_{v}$ amount of the velocity error together with the distance error. The total differential reward is calculated as the sum of all individual link rewards \eqref{eq:differential_reward}.
  \begin{align} \label{eq:differential_reward}
    r_{link} = \sum_{l \in links} w_{d} * (d_{l,t-1} - d_{l,t}) + w_{v} * (v_{l,t-1} - v_{l,t})
  \end{align}
where $d_{l}$ are introduced in \eqref{eq:quat_dist}, \eqref{eq:axis_dist}; $w_{d}$ the link weight coefficients, and $w_{v}$ a small non-negative constant.

  \begin{figure}
    \centering
    \vspace*{1.5mm}
    \includegraphics[width=1.9cm]{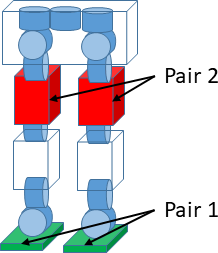}
    \vspace*{-1.5mm}
    \caption{Contact Pairs}
    \label{fig:contacts}
  \end{figure}
  
  The (-$r_{collision}$) term is a penalty for self collisions (Fig. \ref{fig:contacts}). A constant penalty is applied to the reward when labeled contact pairs come into contact.
  
  The $r_{ground}$ term is based on the foot-ground interaction between the agent and world. When processing the motion capture data for training, we add an additional field to each link at each time step indicating if it is on or off the ground. This is used to give a positive reward if the agent foot pressure sensor is above a threshold and the actor's foot recorded as on the ground or alternatively if the sensor is below a threshold and the actor's foot recorded as off the ground \eqref{eq:pressure}.
  \begin{align} \label{eq:pressure}
    \begin{split}
      & r_{P} = \left\{
      \begin{array}{l l}
        r_{P} & if \; gnd_{on} \; \& \; f_{P} > P_{th} \\
        0 & if \; gnd_{on} \; \& \; f_{P} < P_{th} \\
        r_{P} & if \; gnd_{off} \; \& \; f_{P} < P_{th} \\
        0 & if \; gnd_{off} \; \& \; f_{P} > P_{th} \\
      \end{array} \right.
    \end{split}
  \end{align}
  where $gnd_{on}$ and $gnd_{off}$ indicate the foot ground state, $f_{P}$ indicates the foot pressure sensor reading, and $P_{th}$ the pressure threshold.

  An additional term is added to encourage the foot to be parallel to the ground when in contact. It is proportional to the angle between the foot local vertical axis and the ground normal \eqref{eq:projection}.
  \begin{align} \label{eq:projection}
    \begin{split}
      & \vec{e}_z = (0, 0, 1)^T \\
      & \vec{e}_{z,world} = Q_{foot} * \vec{e_z} * Q^{*}_{foot} \\
      & \alpha = cos^{-1}(\vec{e}_{z,world}[z]) \\
      & r_{flat} = K_1 * (K_2 - \alpha)
    \end{split}
  \end{align}
  where $\vec{e}_z$ indicates the vertical axis, $Q_{foot}$ defines the foot orientation quaternion, $\vec{e}_{z,world}$ is the foot up vector in the world reference frame, and $K_1$ and $K_2$ are constants. The complete ground reward is calculated as \eqref{eq:ground}:
  \begin{align} \label{eq:ground}
    \begin{split}
      & r_{ground} = r_{P} + r_{flat}.
    \end{split}
  \end{align}
  
  The (-$r_{limit}$) term provides a penalty on a per joint basis if a target joint position is outside the physical limits of the joint. This penalty pushes the training process to avoid entering areas where the control policy is unable to affect the agent state \eqref{eq:reward_joint_limit}.
  \begin{align}
  \label{eq:reward_joint_limit}
    r_{limit} = \left\{
    \begin{array}{l l}
      k_{joint} * (C_{joint} + (a_i - \alpha^{up}_{i, lim})) & a_i > \alpha^{up}_{i, lim} \\
      k_{joint} * (C_{joint} + (\alpha^{low}_{i, lim} - a_i)) & a_i < \alpha^{low}_{i, lim} \\
    \end{array} \right.
 \end{align}
  where $C_{joint}$ and $k_{joint}$ are constants that define how sharply the penalty increases, $a_{i}$ the $i$-th joint position, $\alpha^{up}_{i, lim}$ the $i$-th joint upper limit and $\alpha^{low}_{i, lim}$ the $i$-th joint lower limit.

\section{TRAINING}
  To speed up training we terminate each rollout early and subtract a penalty from the reward if the agent falls over. Additionally, we early terminate if the distance between any of the agent link orientations and the target link orientations is greater than a specified threshold. Early termination prevents the network from learning while in an unrecoverable state at the cost of fully exploring the state space.

  Peng et al., 2018 \cite{deep_mimic} established the need for starting from random states along a trajectory when training in order to learn complex movements. To achieve this we build up a list of valid start poses during training by comparing the quaternion distances between the mapped agent and target link orientations against a threshold.  At each time step, if all values are below the threshold, we add the pose to a list of start poses.  Existing poses are replaced if the sum of the distances is an improvement over the previous sum.  We sample the starting pose and frame from this list a fixed percentage of the time, beginning from the first frame of the motion when not sampling.

\section{DOMAIN ADAPTATION}
\begin{table}
\vspace*{1.5mm}
\caption{Domain randomization}
\label{tab:randomization}
\begin{center}
\vspace*{-4mm}
\begin{tabular}{|c||c||c|}
\hline
\textbf{Parameter} & \textbf{Lower bound} & \textbf{Upper bound }\\
\hline
Motor backlash & 0 & 5 deg\\
\hline
Friction coef & 0.05 & 0.5\\
\hline
Young's modulus  & $3x10^3$ & $3x10^5$\\
\hline
\end{tabular}
\end{center}
\end{table}

  A real world control system must take into account a number of physical phenomena not present in simulation. To prepare our policy to operate in the real world we adopt the common strategy of domain randomization during training \cite{domain_adaptation}. Our randomization scheme encompasses motor backlash, ground friction, and Young's modulus. Table \ref{tab:randomization} describes the range of values used. 

  We simulate motor backlash as a dead zone around each joint position.  If the position target sent to a joint is within this dead zone, the target is ignored and the joint remains stationary.  Once the position target transitions to outside the dead zone the joint moves as normal.  As a consequence, the total distance the joint must travel is the original distance plus the length needed to overcome the dead zone.  The policy thus learns to overshoot each target position in order to account for backlash in the real world.  We randomize the dead zone between zero and a maximum at the start of each training iteration, increasing the maximum linearly over a configurable number of iterations to gently introduce backlash into the system.

  We randomize friction between a minimum and maximum at the start of each training iteration. This is necessary to prevent the policy from overfitting to the friction model in simulation.
  
  We simulate soft-body surface interactions by randomly selecting the Young's modulus at the start of each training iteration. This teaches the agent to act on soft surfaces where the foot pose may change over time due to the surface deforming under load.

\section{RESULTS}
\begin{figure}
  \centering
  \vspace*{1.5mm}
  \includegraphics[width=7cm]{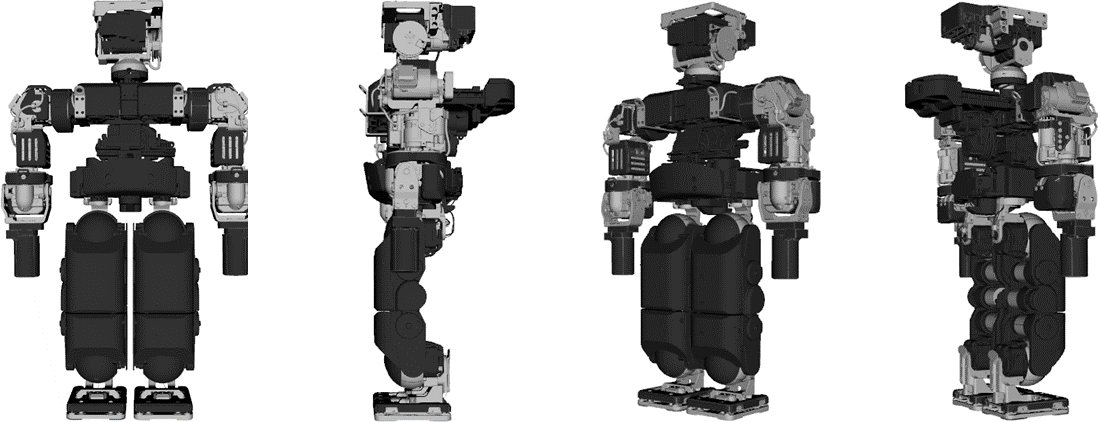}
  \caption{Robot}
  \vspace*{-1.5mm}
  \label{fig:robot}
\end{figure}

\begin{figure}
  \centering
  \vspace*{-1mm}
  \includegraphics[width=7cm]{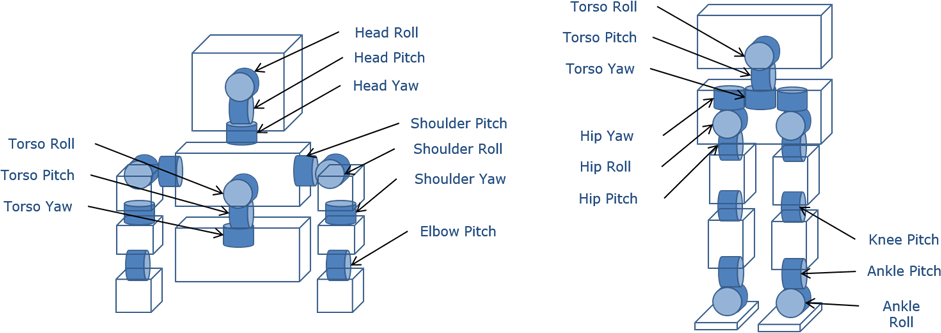}
  \vspace*{-1.5mm}
  \caption{Joints}
  \label{fig:joints}
\end{figure}

Our research was conducted using a robot developed internally for research purposes (Fig. \ref{fig:robot}). The robot has a mass of 1.5 kilograms and a height of 0.3 meters.  It is equipped with four pressure sensors per foot and an IMU sensor attached to the torso.  The robot possesses 26 degrees of freedom (Fig. \ref{fig:joints}).

  We demonstrate our results in simulation using GazeboSim \cite{gazebo}, which utilizes ODE \cite{ode} internally as its physics engine.  We run the simulation at 1024Hz, the control loop at 512Hz, and the policy network at 64Hz.  All neural networks are implemented using Tensorflow \cite{tensorflow}.  Real world results are demonstrated using our internally developed robot.  All motion capture data was obtained using an Optitrack motion capture system \cite{optitrack}.

\begin{figure}
  \centering
  \vspace*{1.5mm}
  \includegraphics[width=8cm]{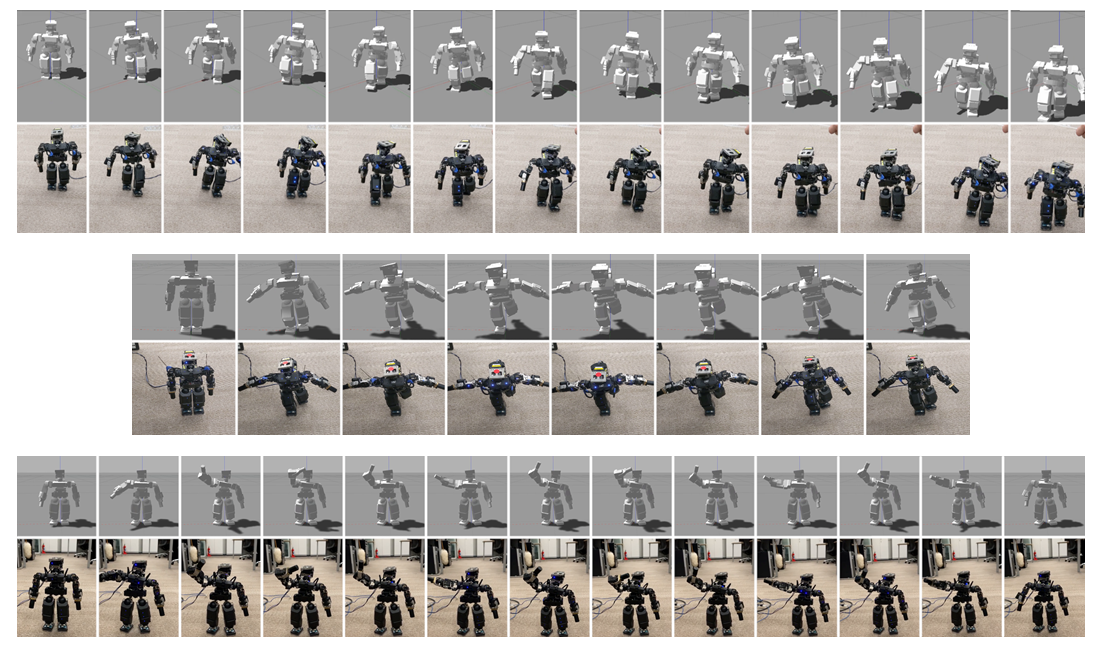}
  \vspace*{-1.5mm}
  \caption{Motion sequences}
  \label{fig:motion_seq}
\end{figure}

  While it is recommended to view the results via the supplementary video, snapshots of the trained movements are available in (Fig. \ref{fig:motion_seq}).  For each learned motion we show the original motion capture data; the robot in simulation trained with backlash, Young's modulus and friction randomization enabled; and the movement performed in the real world using our test robot.  All real world motions are shown performed on two surfaces: a whiteboard representing a hard surface with low friction; and carpet representing a soft-body surface with medium friction.

  A breakdown of the motions is as follows.  The waving motion demonstrates the policies' ability to perform an asymmetric movement as well as to smoothly control the robot's arm.  The two balance motions demonstrate the policies' ability to learn movements that require control over the robots center of mass in order to avoid falling.  The walking motion shows the robot performing a dynamic movement during which it must balance and adjust its stride based on the surface type.  The walk motion is trained to loop indefinitely in order to showcase the robustness of the walk cycle as well as demonstrate control over starting and stopping. The results display our method's ability to inject small visual adjustments into the movements typical for humans.

\section{ABLATIONS}
  We demonstrate the effects of disabling domain randomization during training.  Policies with modified randomization schemes are evaluated using the physical robot on two surfaces: the hard surface white board and soft surface carpet.  The experiments are as follows:

\begin{itemize}
    \item Randomized motor backlash, Young's modulus and dry friction coefficient all disabled.
    \item Randomized motor backlash disabled, randomized dry friction coefficient and Young's modulus enabled.
    \item Randomized Young's modulus disabled, randomized dry friction coefficient and motor backlash enabled.
    \item Randomized dry friction coefficient disabled, randomized Young's modulus and motor backlash enabled.
\end{itemize}

  In order to verify our design decisions we evaluate the following alternative implementations:

\begin{itemize}
  \item Sample from the policy Gaussians at each time step.
  \item Replace the progress based $r_{link}$ reward term with an error based term.
  \item Treat the neural network output as joint positions.
  \item Treat the neural network output as joint accelerations and double integrate.
\end{itemize}
These modified setups are evaluated in simulation due to their high failure rate.

  The real world ablations show that the system failure probability increases when domain randomizations are disabled. This is particularly noticeable for balancing on soft surfaces. The walk movement in contrast is found to be less dependent on randomization. While the walk policy trained without randomizations shows stable behavior early in the trajectory, the chance of failure increases over the course of the movement. Hard surfaces tend to result in more robust movement regardless of the domain randomization scheme used as they closely match the idealized training environment. Soft surfaces prove more difficult and in general require the policy to be trained using domain randomization.  Walking on a soft surface is less dependent on randomization in comparison to balancing.

  The simulation ablations demonstrate: applying random actions at each timestep leads to more conservative movements; an error based reward results in high deviations from the target trajectory, the policy that outputs joint positions repeatedly fails due to high amplitude jitter; the policy that outputs accelerations is unable to learn complicated motions.

\begin{figure}
  \centering
  \vspace*{1.5mm}
  \includegraphics[width=8.5cm]{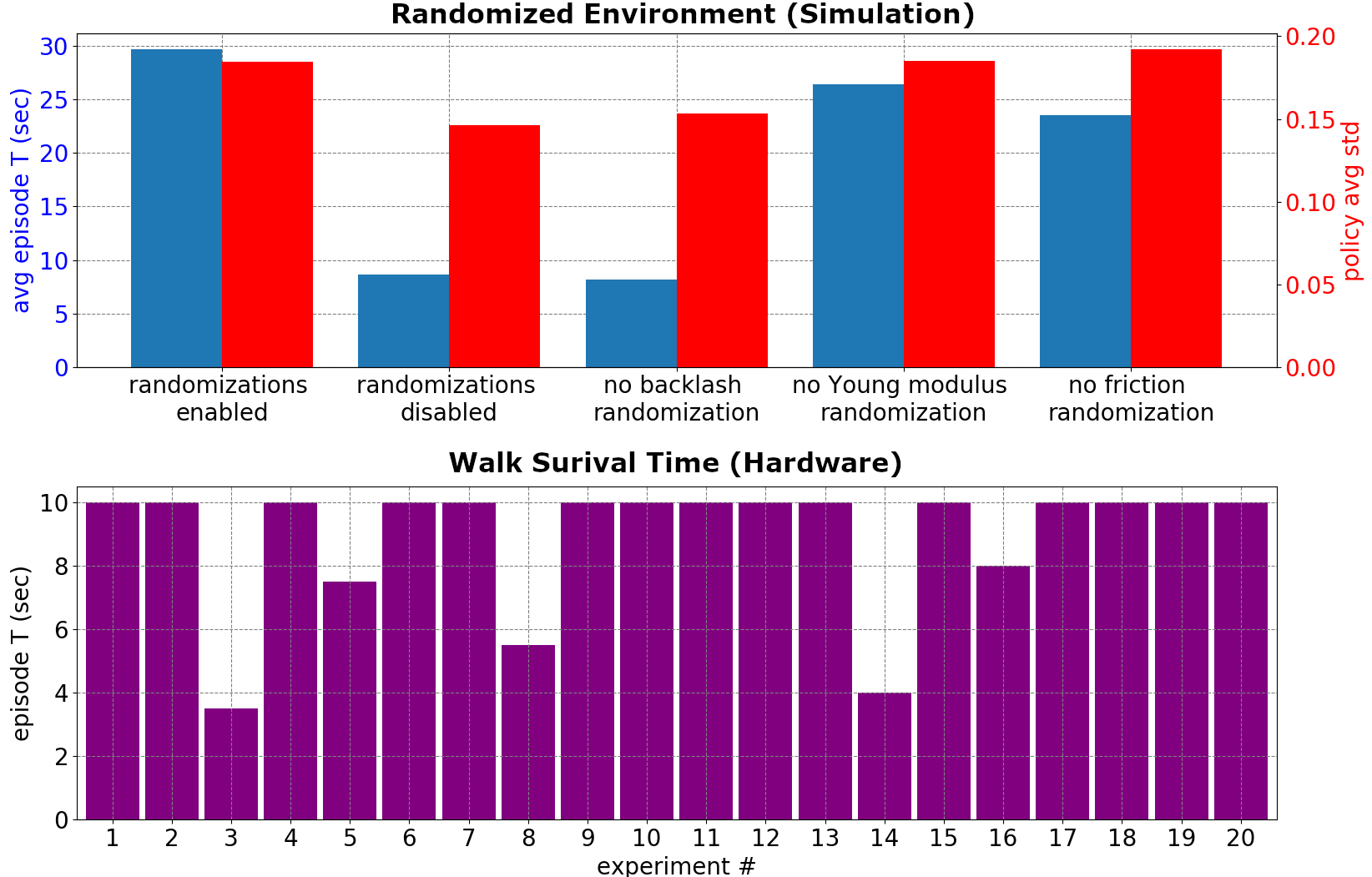}
  \vspace*{-1.5mm}
  \caption{Average episode duration comparison (blue), policy average std comparison (red) and survival time of repeated 10 second walks (purple).}
  \label{fig:ablations_combined}
\end{figure}

  Fig. \ref{fig:ablations_combined} (blue) records the avg. episode duration of the walk motion trained using various domain randomization configurations and tested in a randomized simulation environment. It demonstrates that randomizing all parameters results in a robust policy. Fixing a single parameter results in a decrease in robustness.  Fig. \ref{fig:ablations_combined} (red) examines how the variance of the Gaussian action distribution from the policy varies based on the domain randomization scheme used. Uncertainty in the environment information leads to a larger variance in the action distribution.  Fig. \ref{fig:ablations_combined} (purple) shows the survival time of multiple real world walk attempts performed using the same policy with a backlash randomization from 0 to 2.5 degrees.
  
The ablation footage can be found in the supplementary video. A key takeaway from the footage is that due to the action integration scheme, the policy is less susceptible to producing unsafe erratic behaviors in out-of-distribution states.

\section{CONCLUSION}
  We presented a method for implementing and training a control system capable of performing complex motions on a real world humanoid robot.  Our system facilitates the seamless transition between simulation and real world operation while avoiding the challenges associated with training in the physical world.  The learned motions closely mimic the styles implicit in the target human movements, such as minor adjustments over time made while balancing. The biggest limitation of our method is overall robustness due to static goal selection.  For example, if the robot begins to fall over, it continues to execute the pre-selected motion instead of catching itself. Our next direction is to use physics-based motion matching \cite{motion_matching} in order to perform dynamic goal selection. By performing pose lookups using a combination of the agent's current pose and generated desired future target poses we hope to be able to perform movements such as fall recovery.

\section*{ACKNOWLEDGMENT}
  We would like to express our gratitude to our internal reviewers Xue Bin Peng and Steven Osman for the helpful discussions and feedback, together with Hiroshi Osawa and his team for their development of the robot hardware.

%%%%%%%%%%%%%%%%%%%%%%%%%%%%%%%%%%%%%%%%%%%%%%%%%%%%%%%%%%%%%%%%%%%%%%%%%%%%%%%%

%%%%%%%%%%%%%%%%%%%%%%%%%%%%%%%%%%%%%%%%%%%%%%%%%%%%%%%%%%%%%%%%%%%%%%%%%%%%%%%%

%%%%%%%%%%%%%%%%%%%%%%%%%%%%%%%%%%%%%%%%%%%%%%%%%%%%%%%%%%%%%%%%%%%%%%%%%%%%%%%%
%\section*{APPENDIX}

%Appendixes should appear before the acknowledgment.

%\section*{ACKNOWLEDGMENT}

%The preferred spelling of the word 

%%%%%%%%%%%%%%%%%%%%%%%%%%%%%%%%%%%%%%%%%%%%%%%%%%%%%%%%%%%%%%%%%%%%%%%%%%%%%%%%

%References are important to the reader; therefore, each citation must be complete and correct. If at all possible, references should be commonly available publications.

\clearpage
\appendix

\begin{figure}[hbtp!]
      \centering
      \vspace*{3mm}
      \includegraphics[width=8.6cm]{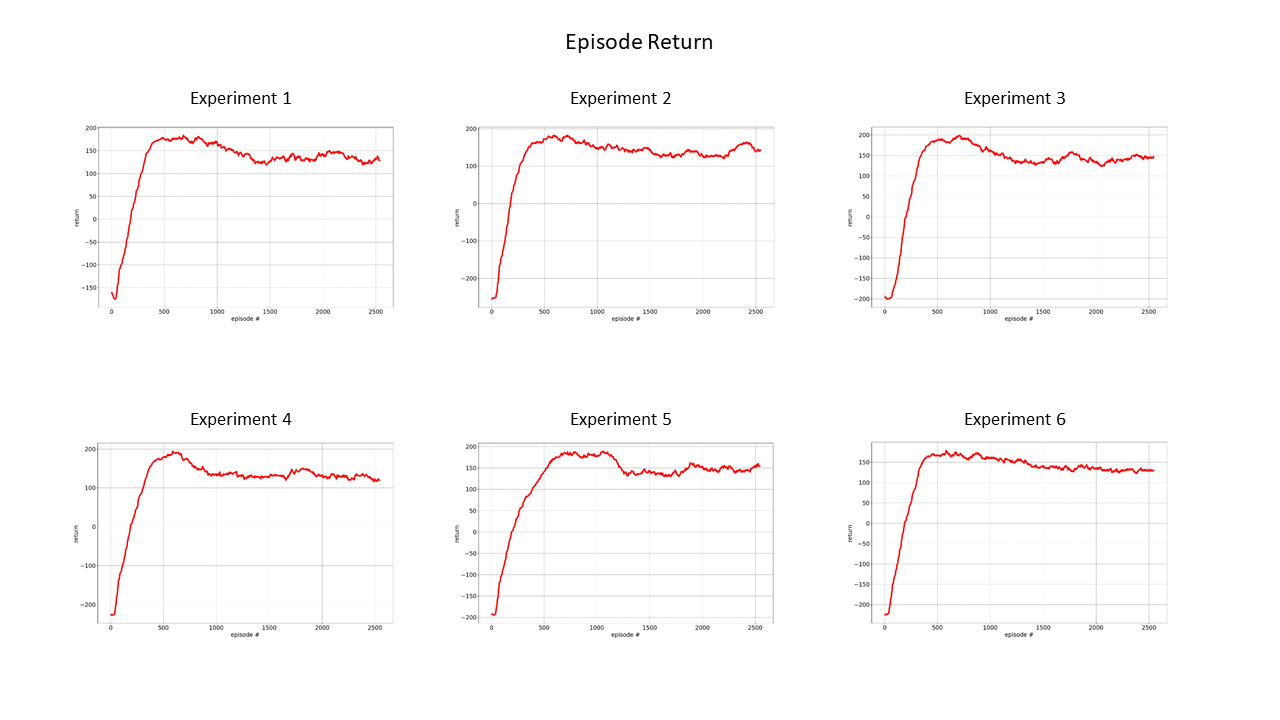}
      \caption{Episode Return.}
      \label{fig:return}
  \end{figure}

\begin{figure}[hbtp!]
      \centering
      \vspace*{3mm}
      \includegraphics[width=8.6cm]{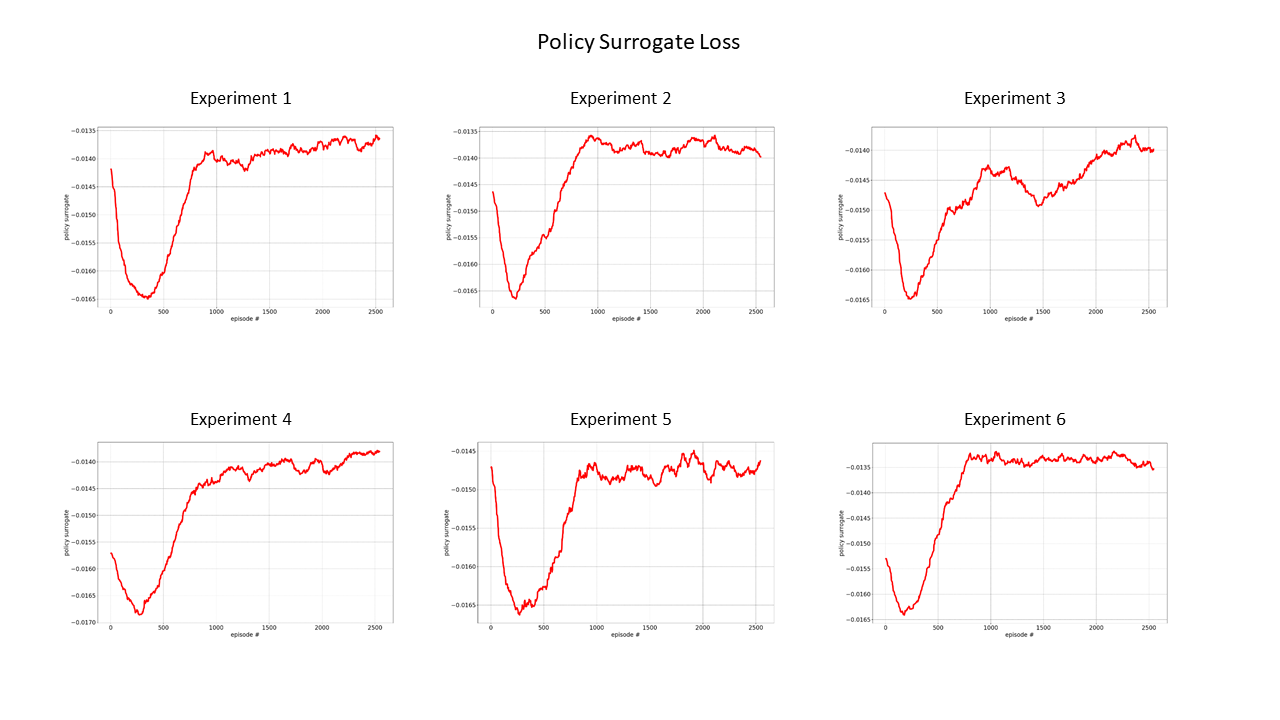}
      \caption{Policy Loss.}
      \label{fig:policy_loss}
  \end{figure}

\begin{figure}[hbtp!]
      \centering
      \vspace*{29mm}
      \includegraphics[width=8.6cm]{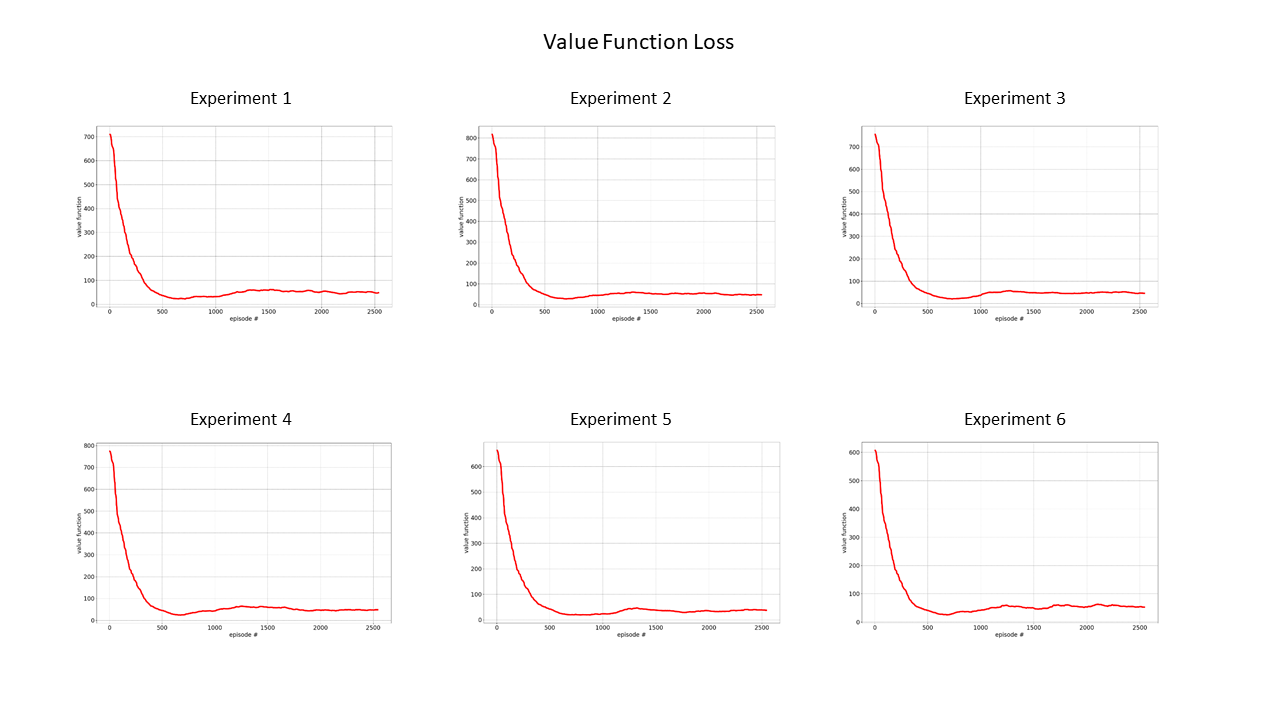}
      \caption{Value Function Loss.}
      \label{fig:vf_loss}
  \end{figure}

\begin{figure}[hbtp!]
      \centering
      \vspace*{3mm}
      \includegraphics[width=8.6cm]{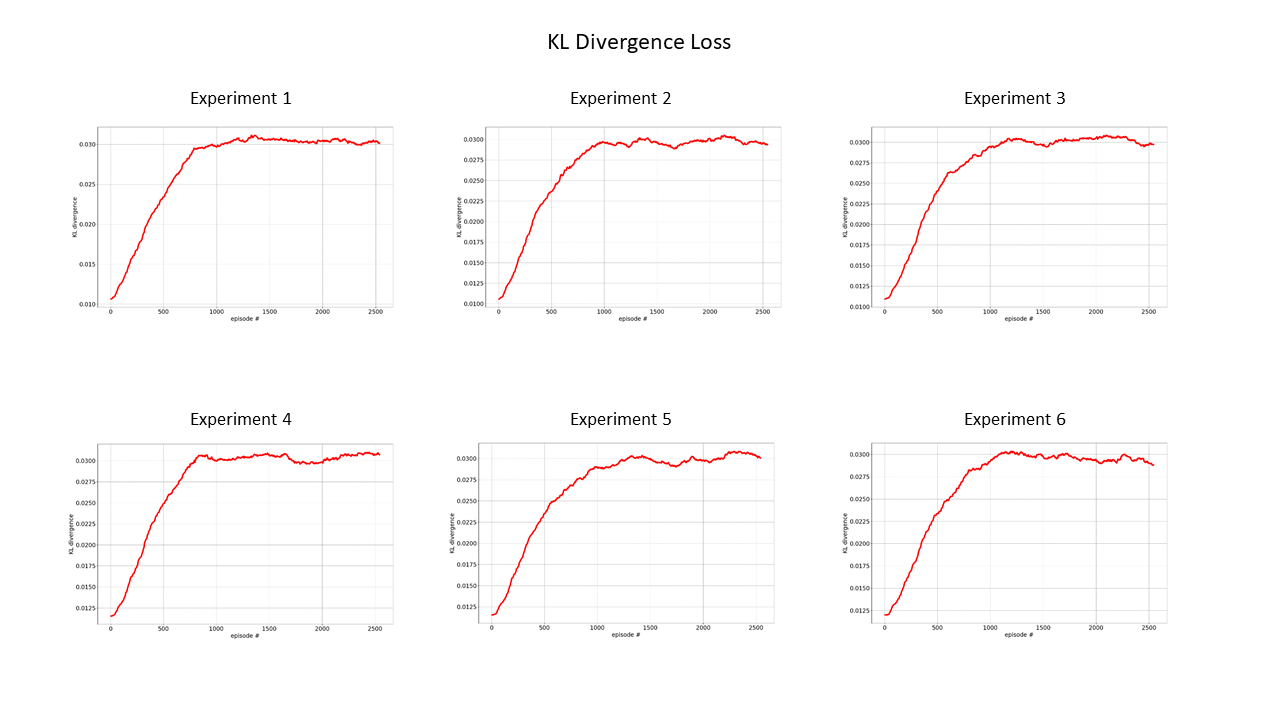}
      \caption{Kullback Leibler Divergence Loss.}
      \label{fig:kl_loss}
  \end{figure}

The graphs (Fig. \ref{fig:return}), (Fig. \ref{fig:policy_loss}), (Fig. \ref{fig:vf_loss}), (Fig. \ref{fig:kl_loss}) show the training results of six experiments initialized with random seeds. They demonstrate training progresses in a similar manner regardless of the starting conditions.

  \newpage

\end{document}